\documentclass[11pt,letterpaper]{article}
\usepackage{emnlp2017}
\usepackage{times}
\usepackage{latexsym}
\usepackage{algorithm}
\usepackage{amsmath}
\usepackage{booktabs}
\usepackage{graphicx}
\usepackage[noend]{algpseudocode}
\usepackage{attrib}

\usepackage{tabularx}
\newcolumntype{L}[1]{>{\raggedright\arraybackslash}p{#1}}
\emnlpfinalcopy

\def\emnlppaperid{***}

\newcommand{\argmax}[1]{\underset{#1}{\operatorname{arg}\,\operatorname{max}}\;}

\title{Revisiting the Centroid-based Method: A Strong Baseline for Multi-Document Summarization}

\author{Demian Gholipour Ghalandari \\
Aylien Ltd., Dublin, Ireland \\
{\tt demian@aylien.com}}
\date{}

\begin{document}

\maketitle

\begin{abstract}
  The centroid-based model for extractive document summarization is a simple and fast baseline that ranks sentences based on their similarity to a centroid vector. In this paper, we apply this ranking to possible summaries instead of sentences and use a simple greedy algorithm to find the best summary. Furthermore, we show possibilities to scale up to larger input document collections by selecting a small number of sentences from each document prior to constructing the summary. Experiments were done on the DUC2004 dataset for multi-document summarization. We observe a higher performance over the original model, on par with more complex state-of-the-art methods. 
\end{abstract}
\section{Introduction}
Extractive multi-document summarization (MDS) aims to summarize a collection of documents by selecting a small number of sentences that represent the original content appropriately. Typical objectives for assembling a summary include information coverage and non-redundancy. A wide variety of methods have been introduced to approach MDS. 

Many approaches are based on sentence ranking, i.e. assigning each sentence a score that indicates how well the sentence summarizes the input ~\cite{erkan2004lexrank, hong2014improving, cao2015ranking}. A summary is created by selecting the top entries of the ranked list of sentences. Since the sentences are often treated separately, these  models might allow redundancy in the summary. Therefore, they are often extended by an anti-redundancy filter while de-queuing ranked sentence lists. 

Other approaches work at summary-level rather than sentence-level and aim to optimize functions of sets of sentences to find good summaries, such as KL-divergence between probability distributions \cite{haghighi2009exploring} or submodular functions that represent coverage, diversity, etc. \cite{lin2011class}

The centroid-based model belongs to the former group: it represents sentences as bag-of-word (BOW) vectors with TF-IDF weighting and uses a centroid of these vectors to represent the whole document collection \cite{radev2004centroid}. The sentences are ranked by their cosine similarity to the centroid vector. This method is often found as a baseline in evaluations where it usually is outperformed ~\cite{erkan2004lexrank, hong2014repository}.  

This baseline can easily be adapted to work at the summary-level instead the sentence level. This is done by representing a summary as the centroid of its sentence vectors and maximizing the similarity between the summary centroid and the centroid of the document collection. A simple greedy algorithm is used to find the best summary under a length constraint.

In order to keep the method efficient, we outline different methods to select a small number of candidate sentences from each document in the input collection before constructing the summary.

We test these modifications on the DUC2004 dataset for multi-document summarization. The results show an improvement of Rouge scores over the original centroid method. The performance is on par with state-of-the-art methods which shows that the similarity between a summary centroid and the input centroid is a well-suited function for global summary optimization.

The summarization approach presented in this paper is fast, unsupervised and simple to implement. Nevertheless, it performs as well as more complex state-of-the-art approaches in terms of Rouge scores on the DUC2004 dataset. It can be used as a strong baseline for future research or as a fast and easy-to-deploy summarization tool. 
  
\section{Approach}
\subsection{Original Centroid-based Method}
The original centroid-based model is described by \citet{radev2004centroid}. It represents sentences as BOW vectors with TF-IDF weighting. The centroid vector is the sum of all sentence vectors and each sentence is scored by the cosine similarity between its vector representation and the centroid vector. Cosine similarity measures how close two vectors $A$ and $B$ are based on their angle and is defined as follows:
\begin{equation}
	sim(A, B) = \frac{A \cdot B}{|A||B|}
\end{equation}
A summary is selected by de-queuing the ranked list of sentences in decreasing order until the desired summary length is reached. 

\citet{Rossiello2017CentroidbasedTS} implement this original model with the following modifications:

\begin{enumerate}
	\item In order to avoid redundant sentences in the summary, a new sentence is only included if it does not exceed a certain maximum similarity to any of the already included sentences. 
	\item To focus on only the most important terms of the input documents, the values in the centroid vector which fall below a tuned threshold are set to zero.
\end{enumerate}
This model, which includes the anti-redundancy filter and the selection of top-ranking features, is treated as the "original" centroid-based model in this paper.

We implement the selection of top-ranking features for both the original and modified models slightly differently to \citet{Rossiello2017CentroidbasedTS}: all words in the vocabulary are ranked by their value in the centroid vector. On a development dataset, a parameter is tuned that defines the proportion of the ranked vocabulary that is represented in the centroid vector and the rest is set to zero. This variant resulted in more stable behavior for different amounts of input documents. 
\subsection{Modified Summary Selection}
The similarity to the centroid vector can also be used to score a summary instead of a sentence. By representing a summary as the sum of its sentence vectors, it can be compared to the centroid, which is different from adding centroid-similarity scores of individual sentences.

With this modification, the summarization task is explicitly modelled as finding a combination of sentences that summarize the input well together instead of finding sentences that summarize the input well independently. This strategy should also be less dependent on anti-redundancy filtering since a combination of redundant sentences is probably less similar to the centroid than a more diverse selection that covers different prevalent topics.

In the experiments, we will therefore call this modification the "global" variant of the centroid model. The same principle is used by the \textit{KLSum} model \cite{haghighi2009exploring} in which the optimal summary minimizes the KL-divergence of the probability distribution of words in the input from the distribution in the summary. \textit{KLSum} uses a greedy algorithm to find the best summary. Starting with an empty summary, the algorithm includes at each iteration the sentence that maximizes the similarity to the centroid when added to the already selected sentences. We also use this algorithm for sentence selection. The procedure is depicted in Algorithm \ref{alg:greedy} below.
\begin{algorithm}[H]
	\begin{algorithmic}[1]
		\State \textbf{Input:} $\text{input sentences } D, \text{centroid } c, limit$
		\State \textbf{Output:} $\text{summary sentences } S$
		\State $S \leftarrow \emptyset$
		\State $length \leftarrow 0$
		\While{$length<limit \textbf{ and } D \ne \emptyset$}  
		\State $s_{best} \leftarrow \argmax{s \in D} sim(S \cup \{s\}, c)$ 
		\State $S \leftarrow S \cup \{s_{best}\}$
		\State $D \leftarrow D \setminus \{s_{best}\}$
		\State $length \leftarrow length + 1$
		\EndWhile\label{greedy}
	\end{algorithmic}
	\caption{Greedy Sentence Selection}
	\label{alg:greedy}
\end{algorithm}
\subsection{Preselection of Sentences}
The modified sentence selection method is less efficient than the orginal method since at each iteration the score of a possible summary has to be computed for all remaining candidate sentences. It may not be noticeable for a small number of input sentences. However, it would have an impact if the amount of input documents was larger, e.g. for the summarization of top-100 search results in document retrieval.

Therefore, we explore different methods for reducing the number of input sentences before applying the greedy sentence selection algorithm to make the model more suited for larger inputs. It is also important to examine how this affects Rouge scores.

 We test the following methods of selecting $N$ sentences from each document as candidates for the greedy sentence selection algorithm:
\subsubsection*{N-first}
The first $N$ sentences of the document are selected. This results in a mixture of a lead-$N$ baseline and the centroid-based method.
\subsubsection*{N-best}
The sentences are ranked separately in each document by their cosine similarity to the centroid vector, in decreasing order. The $N$ best sentences of each document are selected as candidates.
\subsubsection*{New-TF-IDF}
Each sentence is scored by the sum of the TF-IDF scores of the terms that are mentioned in that sentence for the first time in the document. The intuition is that sentences are preferred if they introduce new important information to a document.

Note that in each of these candidate selection methods, the centroid vector is always computed as the sum of all sentence vectors, including the ones of the ignored sentences.
\section{Experiments}
\subsection*{Datasets}
For testing, we use the DUC2004 Task 2 dataset from the Document Understanding Conference (DUC). The dataset consists of 50 document clusters containing 10 documents each. 
For tuning hyperparameters, we use the CNN/Daily Mail dataset \cite{hermann2015teaching} which provides summary bulletpoints for individual news articles.
In order to adapt the dataset for MDS, 50 CNN articles were randomly selected as documents to initialize 50 clusters. For each of these seed articles, 9 articles with the highest word-overlap in the first 3 sentences were added to that cluster. This resulted in 50 documents clusters, each containing 10 topically related articles. The reference summaries for each cluster were created by interleaving the sentences of the article summaries until a length contraint (100 words) was reached.
\subsection*{Baselines \& Evaluation}
 \citet{hong2014repository} published SumRepo, a repository of summaries for the DUC2004 dataset generated by several baseline and state-of-the-art methods \footnote{\url{http://www.cis.upenn.edu/~nlp/corpora/sumrepo.html}}. We evaluate summaries generated by a selection of these methods on the same data that we use for testing. We calculate Rouge scores with the Rouge toolkit \cite{lin2004rouge}. In order to compare our results to \citet{hong2014repository} we use the same Rouge settings as they do\footnote{ROUGE-1.5.5 with the settings -n 4 -m -a -l 100 -x -c 95 -r 1000 -f A -p 0.5 -t 0} and report results for Rouge-1, Rouge-2 and Rouge-4 recall. The baselines include a basic centroid-based model without an anti-redundancy filter and feature reduction.
\subsection*{Preprocessing}
In the summarization methods proposed in this paper, the preprocessing includes sentence segmentation, lowercasing and stopword removal.
\subsection*{Parameter Tuning}
The similarity threshold for avoiding redundancy ($r$) and the vocabulary-included-in-centroid ratio ($v$) are tuned with the original centroid model on our development set. Values from $0$ to $1$ with step size $0.1$ were tested using a grid search. The optimal values for $r$ and $v$ were $0.6$ and $0.1$, respectively. These values were used for all tested variants of the centroid model. For the different methods of choosing $N$ sentences of each document before summarization, we tuned $N$ separately for each, with values from $1$ to $10$, using the global model. The best $N$ found for $N$-first, $N$-best, new-tfidf were 7, 2 and 3 respectively.  
\subsection*{Results}
Table \ref{results} shows the Rouge scores measured in our experiments. 
\begin{table}[t]
	\centering
	\begin{tabular}{l|lll}
		\hline
		Model & R-1 & R-2 & R-4 \\ \hline
		Centroid & 36.03 & 7.89 & 1.20 \\
		LexRank & 35.49 & 7.42 & 0.81 \\
		KLSum & 37.63 & 8.50 & 1.26 \\ \hline
		CLASSY04 & 37.23 & 8.89 & 1.46 \\
		ICSI & 38.02 & \textbf{9.72} & \textbf{1.72} \\
		Submodular & 38.62 & 9.19 & 1.34 \\
		DPP & \textbf{39.41} & 9.57 & 1.56 \\
		RegSum & 38.23 & 9.71 & 1.59 \\ \hline
		Centroid & 37.91 & 9.53 & 1.56 \\
		Centroid + N-first & 38.04 & 9.56 & 1.56 \\
		Centroid + N-best & 37.86 & 9.67 &  \textbf{1.67} \\
		Centroid + new-tf-idf & 38.27 & 9.64 & 1.54 \\
		Centroid + G & 38.55 & 9.73 & 1.53 \\
		Centroid + G + N-first & 38.85 & \textbf{9.86} & 1.62 \\
		Centroid + G + N-best & 38.86 & 9.77 & 1.53 \\
		Centroid + G + new-tf-idf & \textbf{39.11} & 9.81 & 1.58 \\ \hline
		Centroid - R & 35.54 & 8.73 & 1.42 \\
		Centroid + G - R & 38.58 & 9.73 & 1.53
	\end{tabular}
	\caption{Rouge scores on DUC2004.}
	\label{results}
\end{table}
The first two sections show results for baseline and SOTA summaries from SumRepo. The third section shows the summarization variants presented in this paper. "G" indicates that the global greedy algorithm was used instead of sentence-level ranking. In the last section, "- R" indicates that the method was tested without the anti-redundancy filter.

Both the global optimization and the sentence preselection have a positive impact on the performance. 

The global + new-TF-IDF variant outperforms all but the DPP model in Rouge-1 recall. The global + N-first variant outperforms all other models in Rouge-2 recall. However, the Rouge scores of the SOTA methods and the introduced centroid variants are in a very similar range. 

Interestingly, the original centroid-based model, without any of the new modifications introduced in this paper, already shows quite high Rouge scores in comparison to the other baseline methods. This is due to the anti-redundancy filter and the selection of top-ranking features.

In order to see whether the global sentence selection alleviates the need for an anti-redundancy filter, the original method and the global method (without $N$ sentences per document selection) were tested without it (section 4 in Table \ref{results}). In terms of Rouge-1 recall, the original model is clearly very dependent on checking for redundancy when including sentences, while the global variant does not change its performance much without the anti-redundancy filter. This matches the expectation that the globally motivated method handles redundancy implicitly.
\section{Example Summaries}
Table \ref{example} shows generated example summaries using the global centroid method with the three sentence preselection methods. For readability, truncated sentences (due to the 100-word limit) at the end of the summaries are excluded. The original positions of the summary sentences, i.e. the indices of the document and the sentence inside the document are given. As can be seen in the examples, the N-first method is restricted to sentences appearing early in documents. In the new-TF-IDF example, the second and third sentences were preselected because high ranking features such as "robot" and "arm" appeared for the first time in the respective documents. 
\begin{table*}[t]
	\centering
	\begin{tabular}{|p{15.5cm}|}
		\hline
		\textbf{Example Summaries}	
		\\ \hline
		\textbf{N-first} (N=7)\\
		\small
		For the second day in a row, astronauts boarded space shuttle Endeavour on Friday for liftoff on NASA's first space station construction flight. Endeavour and its astronauts closed in Sunday to capture the first piece of the international space station, the Russian-made Zarya control module that had to be connected to the Unity chamber aboard the shuttle. Mission Control gave the astronauts plenty of time for the tasks. On their 12-day flight, Endeavour's astronauts are to locate a Russian part already in orbit, grasp it with the shuttle's robot arm and attach the new U.S. module. 
		
		\textbf{Sentence positions (doc, sent):} (0, 0), (1, 0), (1, 5), (8, 5)
		\\ \hline
		
		\textbf{N-best} (N=2)\\ 
		\small
		For the second day in a row, astronauts boarded space shuttle Endeavour on Friday for liftoff on NASA's first space station construction flight. The astronauts will use the shuttle robot arm to capture the Russian space station piece and attach it to Unity. Mission Control ordered the pilots to fire the shuttle thrusters to put an extra three miles between Endeavour and the space junk, putting Endeavour a total of five miles from the orbiting debris. On their 12-day flight, Endeavour's astronauts are to locate a Russian part already in orbit, grasp it with the shuttle's robot arm and attach the new U.S. module.
		
		\textbf{Sentence positions (doc, sent):} (0, 0), (0, 20), (2, 19), (8, 5)
		\\ \hline
		\textbf{New-TF-IDF }(N=3)\\ 
		\small
		For the second day in a row, astronauts boarded space shuttle Endeavour on Friday for liftoff on NASA's first space station construction flight. The astronauts will use the shuttle robot arm to capture the Russian space station piece and attach it to Unity. The shuttle's 50-foot robot arm had never before been assigned to handle an object as massive as the 44,000-pound Zarya, a power and propulsion module that was launched from Kazakhstan on Nov. 20. Endeavour's astronauts connected the first two building blocks of the international space station on Sunday, creating a seven-story tower in the shuttle cargo bay.

		\textbf{Sentence positions (doc, sent):} (0, 0), (0, 20), (1, 12), (5, 0)
		\\ \hline
	\end{tabular}
	\caption{Summaries of the cluster \textbf{d30031} in DUC2004 generated by the modified centroid method using different sentence preselection methods. }
	\label{example}
\end{table*}
\section{Related Work}
In addition to various works on sophisticated models for multi-document summarization, other experiments have been done showing that simple modifications to the standard baseline methods can perform quite well.

\citet{Rossiello2017CentroidbasedTS} improved the centroid-based method by representing sentences as sums of word embeddings instead of TF-IDF vectors so that semantic relationships between sentences that have no words in common can be captured. 
\citet{mackie2016experiments} also evaluated summaries from SumRepo and did experiments on improving baseline systems such as the centroid-based and the KL-divergence method with different anti-redundancy filters. Their best optimized baseline obtained a performance similar to the ICSI method in SumRepo.
\section{Conclusion}
In this paper we show that simple modifications to the centroid-based method can bring its performance to the same level as state-of-the-art methods on the DUC2004 dataset. The resulting summarization methods are unsupervised, efficient and do not require complicated feature engineering or training.

Changing from a ranking-based method to a global optimization method increases performance and makes the summarizer less dependent on explicitly checking for redundancy. This can be useful for input document collections with differing levels of content diversity.

The presented methods for restricting the input to a maximum of $N$ sentences per document lead to additional improvements while reducing computation effort, if global optimization is being used. These methods could be useful for other summarization models that rely on pairwise similarity computations between all input sentences, or other properties which would slow down summarization of large numbers of input sentences.

The modified methods can also be used as strong baselines for future experiments in multi-document summarization. 
\bibliography{references}

\begin{thebibliography}{}
\expandafter\ifx\csname natexlab\endcsname\relax\def\natexlab#1{#1}\fi

\bibitem[{Cao et~al.(2015)Cao, Wei, Dong, Li, and Zhou}]{cao2015ranking}
Ziqiang Cao, Furu Wei, Li~Dong, Sujian Li, and Ming Zhou. 2015.
\newblock Ranking with recursive neural networks and its application to
  multi-document summarization.
\newblock In {\em AAAI\/}. pages 2153--2159.

\bibitem[{Erkan and Radev(2004)}]{erkan2004lexrank}
G{\"u}nes Erkan and Dragomir~R. Radev. 2004.
\newblock Lexrank: Graph-based lexical centrality as salience in text
  summarization.
\newblock {\em Journal of Artificial Intelligence Research\/} 22:457--479.

\bibitem[{Haghighi and Vanderwende(2009)}]{haghighi2009exploring}
Aria Haghighi and Lucy Vanderwende. 2009.
\newblock Exploring content models for multi-document summarization.
\newblock In {\em Proceedings of Human Language Technologies: The 2009 Annual
  Conference of the North American Chapter of the Association for Computational
  Linguistics\/}. Association for Computational Linguistics, pages 362--370.

\bibitem[{Hermann et~al.(2015)Hermann, Kocisky, Grefenstette, Espeholt, Kay,
  Suleyman, and Blunsom}]{hermann2015teaching}
Karl~Moritz Hermann, Tomas Kocisky, Edward Grefenstette, Lasse Espeholt, Will
  Kay, Mustafa Suleyman, and Phil Blunsom. 2015.
\newblock Teaching machines to read and comprehend.
\newblock In {\em Advances in Neural Information Processing Systems\/}. pages
  1693--1701.

\bibitem[{Hong et~al.(2014)Hong, Conroy, Favre, Kulesza, Lin, and
  Nenkova}]{hong2014repository}
Kai Hong, John~M. Conroy, Benoit Favre, Alex Kulesza, Hui Lin, and Ani Nenkova.
  2014.
\newblock A repository of state of the art and competitive baseline summaries
  for generic news summarization.
\newblock In {\em LREC\/}. pages 1608--1616.

\bibitem[{Hong and Nenkova(2014)}]{hong2014improving}
Kai Hong and Ani Nenkova. 2014.
\newblock Improving the estimation of word importance for news multi-document
  summarization.
\newblock In {\em EACL\/}. pages 712--721.

\bibitem[{Lin(2004)}]{lin2004rouge}
Chin-Yew Lin. 2004.
\newblock Rouge: A package for automatic evaluation of summaries.
\newblock In {\em Text summarization branches out: Proceedings of the ACL-04
  workshop\/}. Barcelona, Spain, volume~8.

\bibitem[{Lin and Bilmes(2011)}]{lin2011class}
Hui Lin and Jeff Bilmes. 2011.
\newblock A class of submodular functions for document summarization.
\newblock In {\em Proceedings of the 49th Annual Meeting of the Association for
  Computational Linguistics: Human Language Technologies-Volume 1\/}.
  Association for Computational Linguistics, pages 510--520.

\bibitem[{Mackie et~al.(2016)Mackie, McCreadie, Macdonald, and
  Ounis}]{mackie2016experiments}
Stuart Mackie, Richard McCreadie, Craig Macdonald, and Iadh Ounis. 2016.
\newblock Experiments in newswire summarisation.
\newblock In {\em European Conference on Information Retrieval\/}. Springer,
  pages 421--435.

\bibitem[{Radev et~al.(2004)Radev, Jing, Sty{\'s}, and Tam}]{radev2004centroid}
Dragomir~R. Radev, Hongyan Jing, Ma{\l}gorzata Sty{\'s}, and Daniel Tam. 2004.
\newblock Centroid-based summarization of multiple documents.
\newblock {\em Information Processing \& Management\/} 40(6):919--938.

\bibitem[{Rossiello et~al.(2017)Rossiello, Basile, and
  Semeraro}]{Rossiello2017CentroidbasedTS}
Gaetano Rossiello, Pierpaolo Basile, and Giovanni Semeraro. 2017.
\newblock Centroid-based text summarization through compositionality of word
  embeddings.
\newblock {\em MultiLing 2017\/} page~12.

\end{thebibliography}
\bibliographystyle{emnlp_natbib}

\end{document}